\documentclass{article} % For LaTeX2e
\usepackage[preprint]{colm2026_conference}

\usepackage{microtype}
\usepackage{hyperref}
\usepackage{url}
\usepackage{booktabs}
\usepackage{amsfonts,amsmath,amssymb,amsthm} 
\usepackage{cleveref}

\usepackage{crossreftools}
\theoremstyle{plain}
\newtheorem{theorem}{Theorem}[section]

\newtheorem{proposition}{Proposition}[section]
\newtheorem{corollary}{Corollary}[section]

\theoremstyle{definition}

\theoremstyle{remark}

\crefname{assumption}{assumption}{assumptions}
\Crefname{assumption}{Assumption}{Assumptions}

\usepackage{dsfont}

\newcommand{\cO}{\mathcal{O}}

\newcommand{\Tr}{\mathop{\text{tr}}\kern.2ex}

\usepackage{xcolor}    % 必须：处理颜色（\colorbox 和 \definecolor）
\usepackage{booktabs}  % 必须：处理表格横线（\toprule, \midrule）
\usepackage{calc}      % 建议：用于计算 \dimexpr 宽度

\definecolor{poscontrib}{HTML}{C0392B} % Base Blue for positive contributions
\definecolor{negcontrib}{HTML}{1F618D} % Base Red for negative contributions
\definecolor{zerocontrib}{HTML}{B0B0B0} % Lighter Gray for zero/neutral highlights

\newcommand{\hlscore}[3]{% % #1: Token/span, #2: Score text, #3: Highlight color string
  \begin{tabular}{@{}c@{}}
    {\setlength{\fboxsep}{1pt}\colorbox{#3}{\strut\color{black}#1}} \\
    \texttt{#2}
  \end{tabular}%
}
\usepackage{fontawesome5}
\usepackage{tabularx}

\usepackage{graphicx} 
\usepackage{subcaption}
\usepackage{fvextra}

\usepackage{lineno}

\definecolor{darkblue}{rgb}{0, 0, 0.5}
\hypersetup{colorlinks=true, citecolor=darkblue, linkcolor=darkblue, urlcolor=darkblue}

\title{ShapE-GRPO: Shapley-Enhanced Reward Allocation for Multi-Candidate LLM Training}

\author{Rui Ai \thanks{MIT. Email: \texttt{ruiai}@mit.edu.}\\
\And
Yu Pan \thanks{University of Sydney. Email: \texttt{yu.pan}@sydney.edu.au.} \\
\And
David Simchi-Levi \thanks{MIT. Email: \texttt{dslevi}@mit.edu.} \\\
\And
Chonghuan Wang \thanks{University of Texas at Dallas. Email: \texttt{chonghuan.wang}@utdallas.edu.}\\
}

\begin{document}

\ifcolmsubmission
\linenumbers
\fi

\maketitle
\begin{abstract}

In user-agent interaction scenarios such as recommendation, brainstorming, and code suggestion, Large Language Models (LLMs) often generate sets of candidate recommendations where the objective is to maximize the collective utility of the entire set rather than individual candidates independently. However, existing reinforcement learning post-training paradigms, such as Group Relative Policy Optimization (GRPO), typically assign the same set-level scalar reward to every candidate in the set. This leads to noisy training signals where poor candidates free-ride on the high reward produced by a single strong peer, resulting in suboptimal exploration. To address this, we propose \underline{Shap}ley-\underline{E}nhanced GRPO (ShapE-GRPO). By leveraging the permutation-invariant nature of set-level utility, 
we derive a Shapley-enhanced formulation from cooperative game theory
% we adapt the Shapley value from cooperative game theory 
to decompose set-level rewards into granular, candidate-specific signals. We show that our formulation preserves the fundamental axioms of the Shapley value while remaining computationally efficient with polynomial-time complexity. Empirically, ShapE-GRPO consistently outperforms standard GRPO across diverse datasets with accelerated convergence during training.

%In many real-world applications, large language models (LLMs) generate multiple candidate recommendations from which users select one to execute, such as choosing a restaurant recommendation. However, existing reinforcement learning post-training paradigms, such as Group Relative Policy Optimization (GRPO), rely on a single sparse reward shared across all candidates. Such a design leads to inefficient reward allocation across candidates and weak training signals at the set level. In this paper, we propose \underline{Shap}ley-\underline{E}nhanced GRPO (ShapE-GRPO), a method that transforms the global sparse reward into a cooperative game among candidate recommendations. We compute candidate-level credit using the Shapley value, enabling efficient reward allocation that does not depend on sequence-level ordering while remaining computationally tractable. Empirically, ShapE-GRPO consistently outperforms standard GRPO across multiple types of datasets with accelerated convergence during training.
\end{abstract}
\section{Introduction}\label{sec:intrp}
In many user–agent interaction scenarios, LLM-based agents generate multiple candidate recommendations from which users select one to execute. Examples include shopping~\citep{li2023gpt4rec}, summarization~\citep{zhang2025systematic}, brainstorming ideas~\citep{lim2024rapid}, and code suggestions~\citep{chen2021evaluating}. In such settings, the objective of generation is no longer to maximize the reward of each individual candidate independently, but rather to maximize the usefulness of the entire set of candidates presented to the user.
For instance, consider a user asking for places to visit near Seattle for a weekend trip. If recommendations are generated solely based on the popularity of individual destinations, an LLM might suggest Mount Rainier, Olympic National Park, and the North Cascades. While each destination is individually popular, this set of recommendations is suboptimal because these locations are highly correlated: a user who is not interested in outdoor activities is unlikely to visit any of them. In contrast, replacing one of these national parks with a destination such as the Boeing Factory Tour, although individually less popular, can significantly improve the coverage of user preferences. As a result, the probability that at least one candidate matches the user’s interests increases. In applications such as shopping recommendations, this directly translates into higher click-through rates and improved downstream value.
% \rui{We need a figure of a chatbox.}
\begin{figure}[!ht]
% \centering % Figure itself can be centered if it's not too wide for the page
\begin{small}
% Using 'l' for the first column (labels) and 'p' for the second to allow content to wrap if necessary,
% but the sequences of scores are designed to be on one line.
% Adjust p column width based on your page layout if sequences become too long.
% \begin{tabular}{@{}lp{\dimexpr0.915\textwidth-2\tabcolsep\relax}@{}} % Adjust width of p column as needed
\begin{tabularx}{\textwidth}{@{}cX@{}}
\toprule
% \textbf{Component} & \textbf{Details} \\
% \midrule
\textbf{\textcolor{red}{\faUser}\ Prompt:} & Please recommend two or three places near Seattle suitable for a trip. \\
% \addlinespace
% % Added <EOS> token to the sequence
% \textbf{Generated Sequence ($y$):} & The plot was amazing, but some acting felt flat. \texttt{<EOS>} \\
% \addlinespace
% \textbf{Overall Reward ($r_\phi(x, y)$):} & $+1.7$ \\
\midrule % Thicker rule to separate sections

% --- Sparse RLHF Reward ---
\textbf{\textcolor{blue}{\faRobot}\ Response:} &
\hlscore{Considering}{+5.0}{poscontrib!28!white}\hspace{0em}%
\hlscore{different}{+5.0}{poscontrib!28!white}\hspace{0em}%
\hlscore{preferences,\,}{+5.0}{poscontrib!28!white}\hspace{-0.8em}%
\hlscore{I\,}{+5.0}{poscontrib!28!white}\hspace{-0.8em}%
\hlscore{recommend\,}{+5.0}{poscontrib!28!white}\hspace{-0.2em}%
\hlscore{the\,}{+5.0}{poscontrib!28!white}\hspace{-0.2em}%
\hlscore{following}{+5.0}{poscontrib!28!white}\hspace{0em}%
\hlscore{locations\,}{+5.0}{poscontrib!28!white}\hspace{-0.1em}%
\hlscore{that\,}{+5.0}{poscontrib!28!white}\hspace{-0.1em}%
\hlscore{cover}{+5.0}{poscontrib!28!white}\hspace{0em}%
\hlscore{natural}{+5.0}{poscontrib!28!white}\hspace{0em}%
\hlscore{scenery,}{+5.0}{poscontrib!28!white}\hspace{0em}%
\hlscore{cultural}{+5.0}{poscontrib!28!white}\hspace{0em}%
\hlscore{landmarks,\,}{+5.0}{poscontrib!28!white}\hspace{-0.1em}%
\hlscore{and\,}{+5.0}{poscontrib!28!white}\hspace{-0.1em}%
\hlscore{technological}{+5.0}{poscontrib!28!white}\hspace{0em}%
\hlscore{marvels:}{+5.0}{poscontrib!28!white}\hspace{0.4em}%
\hlscore{Hiking\,}{+7.5}{poscontrib!45!white}\hspace{-0.4em}%
\hlscore{at\,}{+7.5}{poscontrib!45!white}\hspace{-0.4em}%
\hlscore{Mount}{+7.5}{poscontrib!45!white}\hspace{0em}%
\hlscore{Rainier;}{+7.5}{poscontrib!45!white}\hspace{0.4em}%
\hlscore{Touring\,}{+4.5}{negcontrib!28!white}\hspace{-0.2em}%
\hlscore{the\,}{+4.5}{negcontrib!28!white}\hspace{-0.2em}%
\hlscore{Boeing}{+4.5}{negcontrib!28!white}\hspace{0em}%
\hlscore{Factory;}{+4.5}{negcontrib!28!white}\hspace{0.4em}%
\hlscore{Experiencing}{+3.0}{negcontrib!45!white}\hspace{0em}%
\hlscore{German}{+3.0}{negcontrib!45!white}\hspace{0em}%
\hlscore{culture\,}{+3.0}{negcontrib!45!white}\hspace{-0.4em}%
\hlscore{in\,}{+3.0}{negcontrib!45!white}\hspace{-0.4em}%
\hlscore{Leavenworth.}{+3.0}{negcontrib!45!white}\hspace{0.4em}%
\hlscore{\texttt{<EOS>}}{0.0}{zerocontrib!30!white} \\
% \addlinespace
% \midrule
 
\bottomrule
\end{tabularx}
\end{small}
\caption{When the user considers the ratings for the three suggestions to be 5.0, 4.0, and 3.0, respectively, our ShapE-GRPO performs candidate-level (with broadcasting to token-level) reward allocation as above. However, GRPO only uses a single +5.0 sequence-level reward shared across tokens.}
\label{fig:seattle}
\end{figure}

The core research question of this paper is how to effectively allocate a set-level reward signal to individual candidates within that set to guide more efficient exploration during LLM post-training. Traditional algorithms, such as proximal policy optimization (PPO)~\citep{schulman2017proximal} and GRPO~\citep{shao2024deepseekmath}, typically assign rewards at the level of a single complete response. In scenarios where rewards are determined by a \textit{set} of candidates, these methods assign the \textit{same} scalar reward to every candidate in the set. This approach leads to a significant \textit{reward allocation problem}. For example, in a brainstorming task where a model generates $K$ ideas, a single brilliant suggestion might result in a high set-level reward. Under traditional frameworks, the remaining $K-1$ mediocre or poor ideas would "inherit" that high reward, creating noisy training signals and reinforcing suboptimal behavior. Furthermore, the scale of the state-action space compounds this inefficiency. If a model is capable of generating $N$ distinct ideas and is prompted to produce a set of $K$ candidates, the number of possible combinations grows combinatorially at $\cO(N^K)$. Without a mechanism to provide granular feedback within each set, exploring and learning from such a vast combinatorial space becomes highly sample-inefficient. %For example, in a brainstorming task, suppose that among $K$ generated ideas, there exists one excellent idea along with several poor ones. Because the reward is determined by the overall set outcome, the poor ideas may still receive high rewards simply due to the presence of the excellent one, leading to inefficient and noisy training signals. Moreover, suppose the model is capable of generating $N$ distinct ideas and we prompt it to produce $K$ candidates. The number of possible candidate sets grows combinatorially, resulting in $\cO(N^K)$ possible combinations. Without proper reward allocation within each set, learning from such a large space becomes highly sample-inefficient.

In settings where LLMs act as recommenders generating multiple candidates, the semantic utility and the resulting reward are often \emph{permutation-invariant}, meaning that the total value is independent of the candidates' ordering within the set. This characteristic naturally motivates the use of the Shapley value from cooperative game theory~\citep{shapley1953value} as a principled mechanism for fair reward allocation. Leveraging this framework, we propose an appropriately granular \textit{Shapley-enhanced} reward allocation mechanism. We introduce a candidate-level reward formulation inspired by the Shapley value that decomposes the set-level reward and broadcasts it to the token level. Our approach preserves the fundamental axiomatic properties of the Shapley value: \emph{efficiency}, \emph{symmetry}, \emph{additivity}, and the \emph{null player} property (see \Cref{sec:prop} for details). Furthermore, we demonstrate that under our specific formulation, these exact values can be computed in \emph{polynomial} time, ensuring the method is computationally viable for large-scale LLM post-training. Finally, empirical evaluations across multiple datasets show that our approach significantly enhances training efficiency and model performance compared to the standard GRPO baseline.

\subsection{Our Contributions}
\paragraph{Shapley-Enhanced reward allocation for multi-candidate generation.}
We propose ShapE-GRPO, a Shapley-enhanced reward allocation framework with an appropriate granularity that provides effective candidate-level reward signals for RL post-training in multi-candidate generation settings. It fills the gap in how rewards at the set level can be assigned during LLM post-training, and the underlying candidate-level Shapley values satisfy the standard axioms from cooperative game theory.
%strictly preserving all the desirable properties of Shapley values from cooperative game theory.

\paragraph{Unbiased, informative and efficiently computable candidate-level rewards.}
We theoretically show that under mild assumptions, the proposed candidate-level Shapley rewards are unbiased estimators of the set-level reward and provide meaningful gradients that guide training toward higher-performing candidates. We further prove that in the LLM-as-a-recommender setting, the Shapley values can be computed in polynomial time, ensuring that faster convergence in terms of the number of steps does not come at the cost of increased per-step training time.

\paragraph{Improved performance and a new recommendation benchmark.}
We evaluate ShapE-GRPO on multiple LLM generation tasks, including summarization and code generation, and show that it consistently improves set-level performance compared with the standard GRPO baseline. Furthermore, we release a curated Netflix movie recommendation dataset tailored to the LLM-as-a-recommender scenario, enabling future research on multi-candidate recommendation and set-level optimization for language models.

% {\bf Additional Discussions on Related Works.} 
\subsection{Related Works}
We summarize below three lines of prior work that are most relevant to our study.
\paragraph{RLHF and LLM Alignment}
Many works use reinforcement learning from human feedback (RLHF) to align large language models with human preferences~\citep{christiano2017deep,ouyang2022training,bai2022constitutional,rafailov2023direct,shao2024deepseekmath,yu2025dapo,zheng2025group}, but our work differs in studying multi-candidate generation where utility depends on a set of candidates rather than a single response, and proposing a principled candidate-level reward allocation framework.

\paragraph{Fine-Grained Reward Modeling for LLM Training}
Our work is related to fine-grained reward modeling for LLM training~\citep{chan2024dense,zeng2024token,yoon2024tlcr,xia2024inverse,zhong2024dpo,li2025red,zhou2025t,zhang2025aligndistil,chen2025discriminative,zhu2025tgdpo}, but these methods rely on heuristic token-level decomposition and are inherently order-sensitive. 

\paragraph{Shapley Value in Machine Learning}
Moreover, our work connects to Shapley value in machine learning~\citep{ghorbani2019data,schoch2022cs,liu2023prompt,schoch2023data,wang2024data,horovicz-goldshmidt-2024-tokenshap,cao2025scar,xiao-etal-2025-tokenshapley,tamine2025data}, and differs by formulating reward allocation over unordered candidate sets and preserving computational efficiency. 

We defer further detailed discussion to \Cref{sec:related}.

\section{Preliminaries}
In this section, we briefly review the GRPO objective and the definition of Shapley value from cooperative game theory.

\paragraph{GRPO Algorithm. }
GRPO is a reinforcement learning objective used in LLM post-training that optimizes a policy using group-relative advantages computed from multiple sampled responses. For each prompt $q$, GRPO samples a group of outputs $\{o_1, o_2, \dots, o_G\}$
from the old policy $\pi_{\theta_{\text{old}}}$ and updates the current policy $\pi_\theta$ by maximizing the following objective with an oracle reward function $R(\cdot)$:

\begin{align*}
\mathcal{J}_{\text{GRPO}}(\theta)
&=
\mathbb{E}_{q \sim P(Q),\, \{o_i\}_{i=1}^{G} \sim \pi_{\theta_{\text{old}}}(O \mid q)}
\Bigg[
\frac{1}{G}
\sum_{i=1}^{G}
\frac{1}{|o_i|}
\sum_{t=1}^{|o_i|}
\Big(
\min
\Big(
\frac{\pi_\theta(o_{i,t}\mid q,o_{i,<t})}
{\pi_{\theta_{\text{old}}}(o_{i,t}\mid q,o_{i,<t})}
{A}_{i,t}, \\
&\qquad
\text{clip}\Big(
\frac{\pi_\theta(o_{i,t}\mid q,o_{i,<t})}
{\pi_{\theta_{\text{old}}}(o_{i,t}\mid q,o_{i,<t})},
1-\epsilon, 1+\epsilon
\Big){A}_{i,t}
\Big)
- \beta D_{\mathrm{KL}}\!\left[\pi_\theta \| \pi_{\mathrm{ref}}\right]
\Big)
\Bigg].
\end{align*}

Here ${A}_{i,t}=\frac{R(o_i)-\text{mean}(R)}{\text{std}(R)}$ denotes the advantage shared by every token in the response, $\epsilon$ is the PPO-style clipping parameter, and $\beta$ controls the KL penalty with respect to the reference policy $\pi_{\mathrm{ref}}$. Following prevalent GRPO implementations~\citep{sheng2024hybridflow,nemo-rl}, we employ an augmented standard deviation, clamping $\text{std}(R)$ to 1 when all rewards are identical, to prevent numerical instability.

\paragraph{Shapley Value.}
In cooperative game theory, the Shapley value provides a principled method for allocating the value generated by a coalition of players among individual participants (candidates in our setting). 
Suppose we are given an oracle set-level reward function $R(\cdot)$
mapping any subset of candidates to a non-negative reward. Let the candidate set be $\mathcal{C} = \{c^1, c^2, \dots, c^K\}.
$
The Shapley value assigns a credit $\phi^i$ to each candidate $c^i$ based on its expected marginal contribution to all possible coalitions.
Formally, the Shapley value of candidate $c^i$ is defined as
\begin{align}
\label{eq:shapley}
\phi^i =
\sum_{S \subseteq \mathcal{C} \setminus \{c^i\}}
\frac{|S|!(K-|S|-1)!}{K!}
\Big(
R(S \cup \{c^i\}) - R(S)
\Big)\footnotemark.
\end{align}
Intuitively, the Shapley value measures the expected marginal contribution of candidate $c^i$ when candidates join the coalition in a random order.
\footnotetext{Following the standard convention, we define $R(\emptyset)=0$.}
\section{Methodology}
In this section, we formulate the LLM-as-a-recommender problem as a cooperative game and demonstrate how ShapE-GRPO allocates rewards to individual candidates to provide a correct training signal.
\subsection{Problem Formulation}
For a single interaction between an LLM and a user, we decompose the LLM output $o$ as $o = w + \{c^1, \dots, c^K\}$\footnote{For simplicity, we assume each input generates exactly $K$ outputs. In cases where $K$ varies across inputs, it can be replaced by the respective total number of candidates. Throughout this paper, we use subscripts to denote output indices and superscripts to denote candidate indices.}, where $c^i$ denotes the $i$-th candidate recommendation extracted from the response and $K$ is the total number of candidates. For example, in the travel planning scenario shown in \Cref{fig:seattle}, the candidates correspond to recommended travel destinations. The component $w$ represents the general reasoning or explanation process accompanying the recommendations.

In LLM-as-a-recommender settings, the user reward is typically determined by the best candidate among the generated options. Therefore, under the GRPO framework, the reward of the response is given by
\[
R(o) = R(\{c^1, \dots, c^K\})= \max_{i} R(c^i),
\]
where $R(c^i)$ denotes the utility of the $i$-th candidate, such as whether the user clicks the recommendation or adopts the proposed solution. While our framework supports arbitrary reward functions, we choose $\max$ for its notational simplicity and its alignment with the LLM-as-a-recommender paradigm.

However, this shared reward assignment leads to sparse and misaligned training signals. In particular, when at least one candidate is correct, all candidates within the response may be implicitly treated as successful. As a result, incorrect candidates can free-ride on the success of correct ones, receiving positive reinforcement despite not contributing to the final outcome. This lack of proper incentives can significantly reduce training efficiency.

\subsection{ShapE-GRPO Algorithm}
We view all candidates in a response as players in a cooperative game. Given the candidate set decomposition $\{c^1,\dots,c^K\}$ and the reward function $R(\cdot)$, we compute the scaled Shapley value of each candidate according to \Cref{eq:shapley}, which measures its marginal contribution to the final reward.
For the reasoning component $w$, we retain the original reward assignment
and allocate the sequence-level reward $R(o)$ to all tokens in $w$. In contrast, the candidates receive individualized rewards derived from
their Shapley values.

This design provides several important advantages.
First, the Shapley value distinguishes the marginal contribution of
different candidates to the overall reward, providing a correct and
fine-grained post-training signal.
Second, compared with the token-level Shapley allocation, which depends on the semantic order of tokens, user rewards in recommender scenarios do
not depend on the ordering of recommendations.
Therefore, candidate-level Shapley values preserve all desirable economic properties of the Shapley value.
Finally, although a response may contain thousands of tokens, it usually includes only a small number of candidates. Computing Shapley values at the candidate level therefore introduces only negligible additional computational overhead, which we analyze formally in \Cref{sec:theory}.

Therefore, we introduce our \emph{ShapE-GRPO} algorithm.
With the Shapley values computed for each candidate, we rescale them by the number of candidates $K$ to maintain a comparable reward magnitude.
Note that the original Shapley value decomposes the sequence-level reward
$R(o)$ among candidates. In contrast, ShapE-GRPO redistributes the advantage signal used in policy optimization.
Specifically, let $\phi^j_i$ denote the Shapley value of candidate $c^j_i$ in the $i$-th output.
We define the ShapE-GRPO advantage for the $t$-th token as
\[
A_{i,t}^{\text{ShapE}} = \frac{ R^{\text{ShapE}}(o_{i,t})-\text{mean}(R)}{\text{std}(R)},
\]
where we adopt original $\text{mean}(R)$ and $\text{std}(R)$ implementations in GRPO to avoid extra compute overhead. We define the token-level Shapley-enhanced reward, distinguishing whether
a token belongs to a candidate or to the reasoning part, as
\[
R^{\text{ShapE}}(o_{i,t}) =
\begin{cases}
R(o_i), & \text{if } o_{i,t}\in w_i, \\
K\phi_i^j, & \text{if } o_{i,t}\in c^j_i.
\end{cases}
\]

Using these advantages, the ShapE-GRPO objective becomes
\begin{align*}
\mathcal{J}_{\text{ShapE-GRPO}}(\theta)
&=
\mathbb{E}_{q \sim P(Q),\, \{o_i\}_{i=1}^{G} \sim \pi_{\theta_{\text{old}}}(O \mid q)}
\Bigg[
\frac{1}{G}
\sum_{i=1}^{G}
\frac{1}{|o_i|}
\sum_{t=1}^{|o_i|}
\Big(
\min
\Big(
\frac{\pi_\theta(o_{i,t}\mid q,o_{i,<t})}
{\pi_{\theta_{\text{old}}}(o_{i,t}\mid q,o_{i,<t})}
\textcolor{red}{ A_{i,t}^{\text{ShapE}}}, \\
&\qquad
\text{clip}\Big(
\frac{\pi_\theta(o_{i,t}\mid q,o_{i,<t})}
{\pi_{\theta_{\text{old}}}(o_{i,t}\mid q,o_{i,<t})},
1-\epsilon, 1+\epsilon
\Big)\textcolor{red}{ A_{i,t}^{\text{ShapE}}}
\Big)
- \beta D_{\mathrm{KL}}\!\left[\pi_\theta \| \pi_{\mathrm{ref}}\right]
\Big)
\Bigg].
\end{align*}

By allocating rewards according to each candidate's scaled marginal contribution, ShapE-GRPO eliminates the free-rider problem present in shared reward schemes and provides correct training signals for the
LLM-as-a-recommender setting.
\section{Theoretical Analysis}\label{sec:theory}
In this section, we provide theoretical insights into the proposed
ShapE-GRPO method.
First, we show that when the advantage serves as the directional signal for policy updates, ShapE-GRPO can be interpreted as a reweighting of the
GRPO training signal. This reweighting preserves the overall optimization
structure while eliminating the free-rider problem, ensuring that
incorrect candidates cannot receive positive reward incentives.
Second, we analyze the computational complexity of the proposed
Shapley-enhanced advantage. We show that computing Shapley values at the
candidate level introduces only negligible additional overhead.
This result establishes the efficiency of the candidate-level Shapley
reward allocation scheme.

% \begin{figure}[!ht]
% \centering
% % \begin{subfigure}{0.49\linewidth}
% % \centering
% % \includegraphics[width=\linewidth]{Sections/cv_acl.png}
% % % \caption{GRPO}
% % \end{subfigure}
% % \hfill
% % \begin{subfigure}{0.49\linewidth}
% % \centering
% \includegraphics[width=\linewidth]{Sections/cv_acl.png}
% % \caption{ShapE-GRPO}
% % \end{subfigure}

% \caption{Distribution of candidate length CV for summarization using \textsc{Qwen3-8b}.}
% \label{fig:tokens}
% \end{figure}

\begin{figure}[!t]
\centering

% -------- 左：Figure 1 --------
\begin{minipage}{0.5\linewidth}
    \centering
    \includegraphics[width=\linewidth]{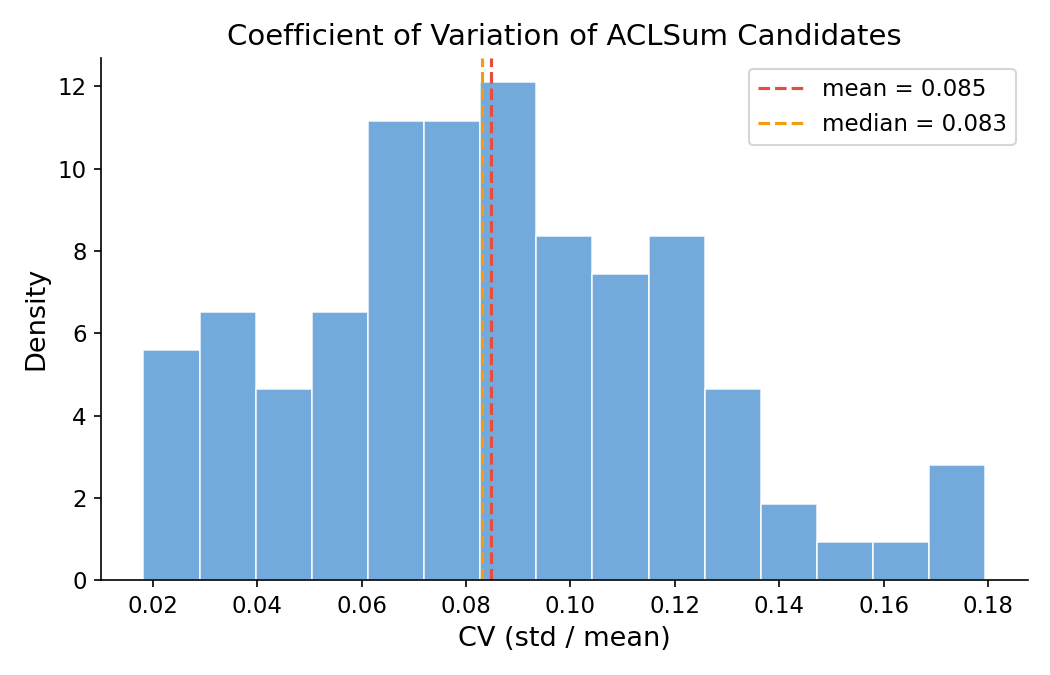}
    \captionof{figure}{Distribution of candidate length CV for summarization.}
    \label{fig:tokens}
\end{minipage}
\hfill
% -------- 右：Figure 2 --------
\begin{minipage}{0.46\linewidth}
    \centering
    \includegraphics[width=\linewidth]{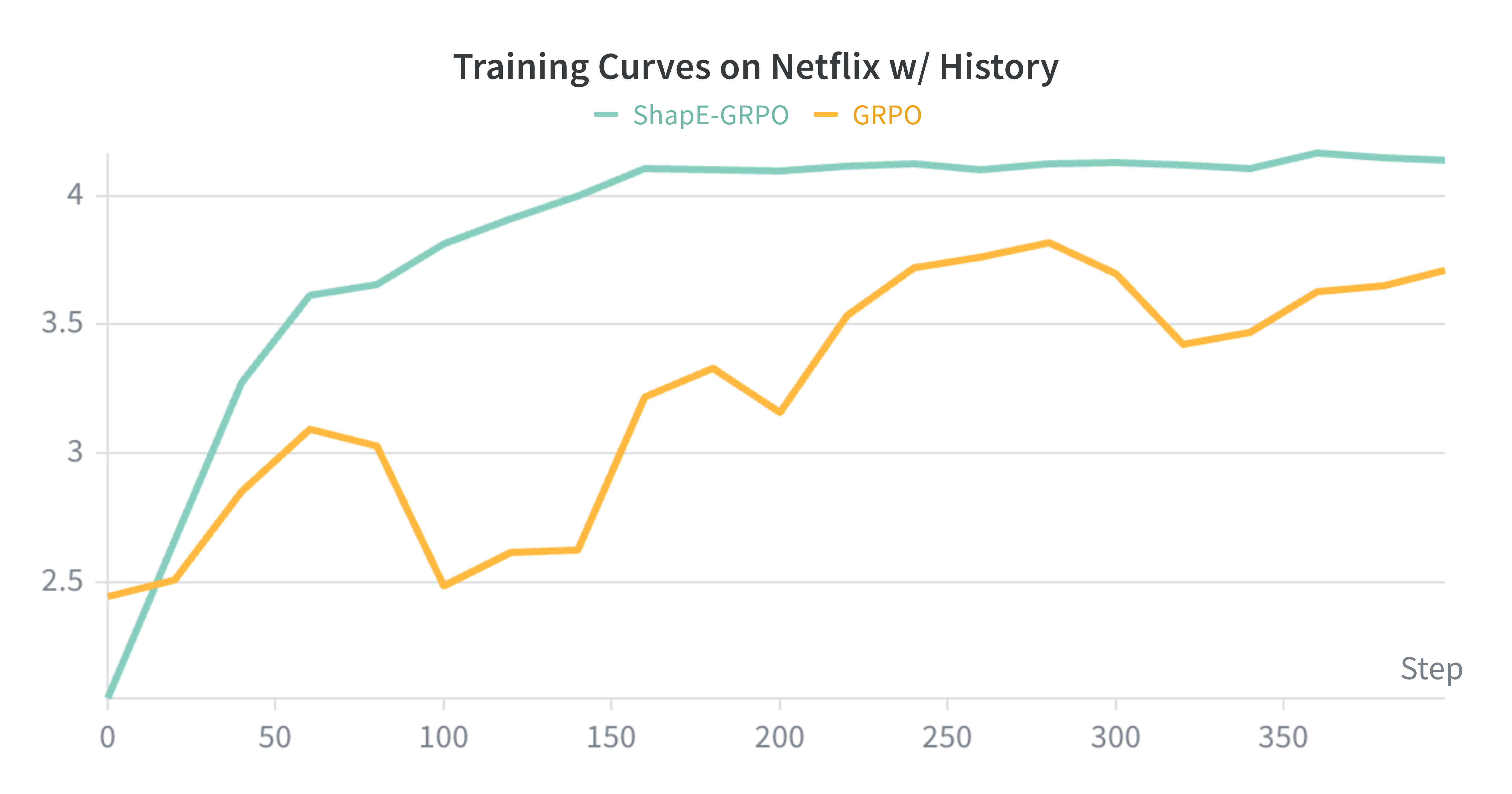}
    
    \captionof{figure}{Training curves on the Netflix dataset with user history.
    ShapE-GRPO converges faster and exhibits significantly more stable training dynamics compared with GRPO.}
    \label{fig:training_curve}
\end{minipage}

\end{figure}

We now consider an output $o_i$. Unlike different outputs, which may have
large variations in length, the number of tokens associated with different
candidates is typically similar in practice. For example, in travel planning
tasks, candidates correspond to location names, while in movie recommender
systems, candidates may be represented by fixed-length movie identifiers.
Therefore, the token counts of different candidates are approximately
balanced. 
We present in \Cref{fig:tokens} the distribution of the coefficient of variation (CV) of candidate lengths using \textsc{Qwen3-8b}~\citep{yang2025qwen3}, namely token numbers, for the textual summarization task used in our experiments. The variation is significantly smaller than the response length variation reported in prior work~\citep{perez2025castillo}.
Based on this observation, we present the following proposition showing
that ShapE-GRPO does not introduce additional bias compared with the
original GRPO objective.

\begin{proposition}\label{prop:unbias}
Assuming that the candidates $\{c_i^1,\dots,c_i^K\}$ in the output $o_i$
have equal lengths, the ShapE-GRPO advantage is a reweighting of the
GRPO advantage. Specifically,
\[
\sum_{t=1}^{|o_i|} A_{i,t}
=
\sum_{t=1}^{|o_i|} A^{\text{ShapE}}_{i,t}.
\]
\end{proposition}

Moreover, ShapE-GRPO ensures that each candidate receives an appropriate
reward allocation according to its marginal contribution, thereby
eliminating the free-rider phenomenon. In particular, incorrect candidates
cannot obtain positive reward signals due to the success of other
candidates. This prevents false-positive reinforcement in the RL training
process and can lead to faster convergence.
\begin{proposition}\label{prop:pos}
Consider the candidates $\{c_i^1,\dots,c_i^K\}$ in the output $o_i$ and a general non-negative reward function $R(\cdot)$. Then, for any incorrect candidate $c_i^j$ with $R(c_i^j)=0$, it holds that
% sorted in descending order of their rewards, i.e.,
% \[
% R(c_i^{(1)}) \ge \cdots \ge R(c_i^{(K)}).
% \]
the ShapE-GRPO
advantage satisfies
\[
A_{i,t}^{\text{ShapE}} \le 0 \text{ for any token $o_{i,t} \in c_i^{j}$}.
\]
\end{proposition}
We note that our prompt engineering explicitly instructs the LLM to
generate $K$ distinct candidates in each response. This design provides
denser reward signals for training. At the same time, it partially
mitigates an issue observed in standard GRPO training, where repeated
sampling may produce identical responses, leading to degenerate reward
variance and vanishing advantage signals~\citep{zhang2025edge}.
By encouraging diversity among candidates within a single response, our framework improves the efficiency of reward feedback and stabilizes
policy updates. Extending this idea to the general GRPO setting,
even when only a single candidate is required to be generated per response, may
constitute a promising direction for future research.

Finally, we show that our algorithm incurs only negligible additional
overhead compared with GRPO. Specifically, due to the special
structure of the LLM-as-a-recommender setting, the exact Shapley values
can be computed with only $\mathcal{O}(K^2)$ additional floating-point
operations, rather than the exponential $\mathcal{O}(2^K)$ complexity
in the general case.
\begin{theorem}\label{thm:sort}
Consider the candidates $\{c_i^1,\dots,c_i^K\}$ in the output $o_i$,
sorted in descending order of their rewards, i.e.,
\[
R(c_i^{(1)}) \ge \cdots \ge R(c_i^{(K)}).
\]
Then for any $o_{i,t}\in c_i^{(j)}$, its Shapley-enhanced reward 
% associated with a general non-negative reward function $R(\cdot)$ 
is given by
\begin{equation}\label{eq:order}
   R^{\text{ShapE}}(o_{i,t}) =
\sum_{k=j}^{K}
\frac{K(R(c_i^{(k)}) - R(c_i^{(k+1)}))}{k}, 
\end{equation}
where we define $R(c_i^{(K+1)}) = 0$.

In particular, in the binary reward case where
$R(\cdot) \in \{0,1\}$, suppose there are $m \ge 1$
correct candidates. Then, the tokens within these candidates receive a Shapley-enhanced reward $\frac{K}{m}$,
while all incorrect candidates receive zero reward.
\end{theorem}
This immediately leads to the following corollary, which shows that
ShapE-GRPO introduces only polynomial additional computational overhead.
\begin{corollary}\label{corr:flop}
For a general reward function $R(\cdot)$, computing the
ShapE reward $R^{\text{ShapE}}(\cdot)$ for each output requires only an
additional $\mathcal{O}(K^2)$ floating-point operations.
In the binary reward case, the complexity further reduces to
$\mathcal{O}(K)$.
\end{corollary}
In summary, our setting allows the Shapley values to be computed efficiently and exactly. Unlike many machine learning applications, where Shapley values must be approximated, our algorithm admits an
exact computation in $\mathcal{O}(K^2)$ time.
Moreover, in contrast to token-wise Shapley allocation, our method only computes Shapley values over the $K$ candidates. In typical LLM scenarios, the number of candidates is significantly smaller
than the number of tokens in a response. By fixing the reward of the reasoning component to $R(o)$, we avoid introducing reward bias while also eliminating the dependence on token order. This yields
an appropriately grained reward allocation scheme for the
LLM-as-a-recommender setting.
Finally, we note that the Shapley-enhanced reward only requires
floating-point computations. Since gradients of the advantage
are not needed during training, the method avoids additional
gradient computations entirely. These three properties ensure
that ShapE-GRPO introduces negligible computational overhead.
At the same time, the denser reward allocation improves the
training signal, which empirically accelerates convergence
toward the targeted policy and often leads to faster training.

\section{Experiments}
%\rui{To Yu: please think about what figures we can add, considering what we have done. }
In this section, we evaluate ShapE-GRPO on multiple tasks,
including summarization, code suggestion, and movie
recommendation. Across all evaluated datasets, ShapE-GRPO
consistently outperforms the standard GRPO baseline,
demonstrating the effectiveness of the proposed reward
allocation scheme.
% \begin{figure}[!t]
% \centering

% % -------- 左：Figure 1 --------
% \begin{minipage}{0.28\linewidth}
%     \centering
%     \includegraphics[width=\linewidth]{Sections/cv_acl.png}
%     \captionof{figure}{Distribution of candidate length CV for summarization.}
%     \label{fig:tokens}
% \end{minipage}
% \hfill
% % -------- 右：Figure 2 --------
% \begin{minipage}{0.70\linewidth}
%     \centering
%     \includegraphics[width=0.49\linewidth]{Sections/GRPO.png}
%     \hfill
%     \includegraphics[width=0.49\linewidth]{Sections/ShapE.png}
    
%     \captionof{figure}{Training curves on the Netflix dataset with user history.
%     ShapE-GRPO converges faster and exhibits significantly more stable training dynamics compared with GRPO.}
%     \label{fig:training_curve}
% \end{minipage}

% \end{figure}

\subsection{Experimental Setup}

% \paragraph{Evaluation Tasks and Models.}
We evaluate ShapE-GRPO on three representative tasks. We train the \textsc{Qwen3-8B} model~\citep{yang2025qwen3} for all experiments, using two GH200 GPUs.
Additional implementation details, e.g., prompts and hyperparameter settings,
as well as results with \textsc{Llama3.1-8B-Instruct}~\citep{grattafiori2024llama}, 
are provided in \Cref{app:exp}.

\paragraph{Summarization.} 
We use the ACLSum dataset~\citep{takeshita2024aclsum}, a high-quality benchmark for
scientific paper summarization. ACLSum contains scientific
publications from major NLP venues and provides expert-annotated
summaries that capture multiple aspects of each paper,
including the challenge, approach, and
outcome of the work. The dataset is manually curated
and validated by domain experts to ensure high-quality
reference summaries.
For each input document, the model generates multiple candidate
summaries. We adopt an LLM-as-a-judge evaluation protocol~\citep{zheng2023judging} and
use \textsc{GPT-5.2}~\citep{singh2025openai} to score the generated summaries based on
their quality relative to the reference summaries. The obtained
scores are then normalized to the continuous range $[0,1]$ and used as the
reward signal during training.

\paragraph{Code Suggestion.} 
We use the DS-1000 dataset, a widely used benchmark for
data science code generation~\citep{lai2023ds}. DS-1000 contains programming problems covering practical data science
tasks across several popular Python libraries, including
\texttt{NumPy}, \texttt{Pandas}, \texttt{SciPy},
\texttt{Scikit-learn}, \texttt{Matplotlib},
\texttt{TensorFlow}, and \texttt{PyTorch}. The problems
are constructed from real-world StackOverflow questions
and reformulated to reflect realistic programming scenarios.
To simulate a typical user interaction where a programmer
asks an LLM for coding suggestions, the model generates multiple candidate code solutions for each prompt.
The user only needs one working solution among the
suggestions to solve the problem. Therefore, we adopt
a binary reward scheme: the reward is $1$ if at least one
candidate solution passes the DS-1000 execution tests,
and $0$ otherwise. This setup naturally matches the LLM-as-a-recommender formulation studied in this paper.

\paragraph{Movie Recommendation.} 
Existing benchmarks for LLM-as-a-recommender are limited.
To evaluate our method in a realistic recommendation setting,
we construct a dataset based on the Netflix Prize data\footnote{Original data can be found at \url{https://www.kaggle.com/datasets/netflix-inc/netflix-prize-data}}.
We tailor the dataset to simulate a video platform
recommendation scenario. For each user, we use their ratings
for movies released before 2005 as the user history and ask
the model to recommend movies released in 2005.
This temporal split prevents information leakage between
training signals and evaluation targets.
To study the impact of user context, we provide two versions
of the task: one with user history and one without user
history. The model generates multiple candidate movie recommendations for each user, mimicking the behavior
of real-world video platforms where multiple suggestions
are displayed after a user finishes watching a video.
We use the user's ground-truth rating for the recommended
movie as the reward signal, following the original
five-point rating scale in the Netflix dataset.
The objective is therefore to maximize the expected user
rating of the recommended movies, which can be interpreted
as a proxy for user engagement or click-through rate
in real-world recommendation systems.

\subsection{Main Results}
% ShapE-GRPO consistently improves the performance of the
% pretrained model during post-training across multiple datasets, including summarization, code suggestion, and movie recommendation tasks.
% The key difference between ShapE-GRPO and GRPO lies in the training signal: ShapE-GRPO replaces the standard advantage $A_{i,t}$ in GRPO with the Shapley-enhanced advantage $A^{\text{ShapE}}_{i,t}$, which allocates rewards at the candidate level.
% To study the impact of this design, we conduct an ablation analysis on the proposed candidate-level reward allocation.
% The results show that this appropriately grained and denser reward signal consistently outperforms the shared reward
% scheme used in GRPO in the LLM-as-a-recommender setting.
ShapE-GRPO consistently improves the performance of the
pretrained model during post-training across multiple datasets, including summarization, code suggestion, and movie recommendation tasks.
The key difference between ShapE-GRPO and GRPO lies in the training signal: ShapE-GRPO replaces the standard advantage $A_{i,t}$ in GRPO with the Shapley-enhanced advantage $A^{\text{ShapE}}_{i,t}$, which allocates rewards at the candidate level.
To study the impact of candidate-level reward allocation, we additionally compare against a Winner-Takes-All (WTA) baseline.
% WTA also adopts a candidate-level reward allocation, but only assigns reward to the highest-scoring candidate within a response and zero rewards to the remaining candidates, while keeping the reasoning-part reward and the outer GRPO optimization unchanged.
WTA adopts a candidate-level reward allocation that assigns reward only to the highest-scoring candidate within each response, with all remaining candidates receiving zero, while keeping the reasoning-part reward and the outer GRPO optimization unchanged.
When multiple candidates achieve the same highest score, we split the candidate-level reward evenly among them.
This comparison tests whether the best-only reward is already sufficient, or whether coalition-aware reward allocation is needed.

% \begin{table}[!ht]
% \centering
% \small
% \begin{tabular}{lccc}
% \toprule
% Dataset & \textsc{Qwen3-8B} (Pretrained) & GRPO & ShapE-GRPO \\
% \midrule
% ACLSum ($\uparrow$) & 0.654 & 0.658 & \textbf{0.672} \\
% DS-1000 ($\uparrow$) & 0.398\footnotemark & 0.465 & \textbf{0.510} \\
% Netflix w/ history ($\uparrow$) & 2.872\footnotemark & 3.710 & \textbf{4.141} \\
% Netflix w/o history ($\uparrow$) & 3.395 & 3.627 & \textbf{3.821} \\
% \bottomrule
% \end{tabular}
% \caption{Performance comparison across four datasets.
% ShapE-GRPO consistently improves over both the pretrained
% model and the standard GRPO baseline.}
% \label{tab:main_results}
% \end{table}
\begin{table}[!ht]
\centering
\small
\setlength{\tabcolsep}{4pt}
\begin{tabular}{lcccc}
\toprule
Dataset & \shortstack{\textsc{Qwen3-8B}} & GRPO & \shortstack{WTA} & ShapE-GRPO \\
\midrule
ACLSum ($\uparrow$) & 0.654 & 0.658 & 0.642 & \textbf{0.672} \\
DS-1000 ($\uparrow$) & 0.398\footnotemark & 0.465 & 0.418 & \textbf{0.510} \\
Netflix w/ history ($\uparrow$) & 2.872\footnotemark & 3.710 & 3.550 & \textbf{4.141} \\
Netflix w/o history ($\uparrow$) & 3.395 & 3.627 & 2.497 & \textbf{3.821} \\
\bottomrule
\end{tabular}
\caption{Performance comparison across four datasets using \textsc{Qwen3-8B}.
Winner-Takes-All assigns candidate-level reward only to the highest-scoring candidate, whereas ShapE-GRPO allocates reward by scaled Shapley values.
ShapE-GRPO consistently improves over both the standard GRPO baseline and the Winner-Takes-All baseline.}
\label{tab:main_results}
\end{table}
\footnotetext{We limit the maximum output length to 4096 tokens during training.
Since the pretrained model often exceeds this limit, we allow up to 16384 tokens when evaluating the pretrained model.
When restricted to 4096 tokens, the pretrained model achieves a
score of 0.273.}
\footnotetext{When restricted to 4096 tokens, the pretrained model achieves a score of 2.434.}
% winner-takes-all (qwen3):
% ACLSum final reward 0.642
% Netflix w/ hist final reward 1.3277/3.54968 (cf. 644867.out)
% Netflix w/o hist final reward 2.4968
% DS-1000 final reward 0.4175

% \Cref{tab:main_results} summarizes the main results.
% Across all datasets, ShapE-GRPO consistently improves performance over the standard GRPO baseline.
% On ACLSum, ShapE-GRPO increases the score from 0.658 to 0.672, an absolute improvement of 0.014.
% On DS-1000, the performance improves from 0.465 to 0.510,
% corresponding to a relative improvement of 9.7\%.
% The gains are even larger in the recommendation setting:
% on Netflix with user history, ShapE-GRPO improves the average rating from 3.710 to 4.141, a substantial gain of 0.431 points. Similarly, on Netflix without user history, the score increases from 3.627 to 3.821.
% Notably, the \textsc{Qwen3-8B} model struggles to effectively utilize user history. After ShapE-GRPO training, however, the model learns to better incorporate historical preferences, resulting in a much larger performance gain when user history is available.
\Cref{tab:main_results} summarizes the main results.
Across all datasets, ShapE-GRPO consistently improves over both the standard GRPO baseline and the Winner-Takes-All baseline.
This shows that the gains are not explained by a heuristic that only rewards the single best candidate within each response.
Instead, the results support the value of coalition-aware candidate reward allocation under the set-level objective.

On ACLSum, ShapE-GRPO improves over both GRPO and Winner-Takes-All, indicating that the best-only reward is too myopic even for continuous summary-quality rewards.
On DS-1000, ShapE-GRPO again achieves the strongest performance.
The gains are even larger in the recommendation setting: on Netflix with user history, ShapE-GRPO outperforms both baselines by a wide margin, and on Netflix without user history it still yields the best average rating.
Notably, the \textsc{Qwen3-8B} model struggles to effectively utilize user history.
After ShapE-GRPO training, however, the model learns to better incorporate historical preferences, resulting in a much larger performance gain when user history is available.

We further compare the training dynamics of ShapE-GRPO and GRPO on the Netflix dataset with user history. This is the most challenging dataset in our experiments: the average prompt length is around fifteen thousand tokens, and it most closely reflects the intended LLM-as-a-recommender scenario.

% \begin{figure}[!ht]
% \centering
% \begin{subfigure}{0.49\linewidth}
% \centering
% \includegraphics[width=\linewidth]{Sections/GRPO.png}
% % \caption{GRPO}
% \end{subfigure}
% \hfill
% \begin{subfigure}{0.49\linewidth}
% \centering
% \includegraphics[width=\linewidth]{Sections/ShapE.png}
% % \caption{ShapE-GRPO}
% \end{subfigure}

% \caption{Training curves on the Netflix dataset with user history.
% ShapE-GRPO converges faster and exhibits significantly more stable
% training dynamics compared with GRPO.}
% \label{fig:training_curve}
% \end{figure}

\Cref{fig:training_curve} shows the training score
curves of the two methods. ShapE-GRPO reaches saturation
after approximately 160 training steps, after which the
performance remains nearly flat. In contrast, GRPO requires around 240 steps to reach its final performance level.
Moreover, the checkpoints produced during GRPO training
exhibit large fluctuations in evaluation performance.
Specifically, we evaluate the model every 20 training steps and observe large fluctuations in evaluation-set performance during GRPO training. This behavior reflects the inherently high variance caused by sparse reward signals in GRPO. In comparison, ShapE-GRPO produces a much more stable training trajectory due to the denser and more informative reward allocation.

These consistent improvements across diverse tasks and faster convergence rates suggest that more accurate reward allocation at the candidate level leads to more effective and efficient policy optimization.
By allocating rewards according to each candidate's rescaled marginal contribution, ShapE-GRPO provides a more informative training signal than both the shared reward scheme used in GRPO and the heuristic best-only reward used in Winner-Takes-All.
\section{Conclusion and Discussion}
In this paper, we propose ShapE-GRPO, a Shapley-enhanced
policy optimization framework for the LLM-as-a-recommender setting. We show both theoretically and empirically that ShapE-GRPO consistently improves the performance of the standard GRPO algorithm across different model architectures and tasks.
Our key insight is that reward signals in LLM-as-a-recommender problems are naturally generated at the set level.
By allocating rewards according to each candidate's marginal
contribution, ShapE-GRPO provides a permutation-invariant, effective and efficient reward allocation scheme. The proposed Shapley-enhanced advantage preserves the original motivation of Shapley value while remaining computationally tractable in the candidate-level setting.
Together, these results demonstrate that candidate-level reward allocation provides a principled and practical solution for reinforcement learning in LLM-as-a-recommender scenarios.

Questions naturally arise for future exploration. While Shapley value provides a principled way to allocate rewards among candidates, it is only one possible credit allocation scheme. Recent proposals of alternative value concepts, such as instrumental value~\citep{ai2024instrumental}, raise the question of how different
allocation mechanisms may affect learning dynamics and
policy performance. In addition, generating multiple
candidates within a single response appears to increase
output diversity, which may further improve the efficiency
and diversity of GRPO sampling. Another interesting
direction concerns the scoring of candidates: when an
oracle reward model is unavailable, it remains unclear how unsupervised approaches, e.g., gradient-based~\citep{jung2025prismatic}, would perform for candidate evaluation. We leave these
interesting questions as promising directions for
future work.

\bibliography{colm2026_conference}
\bibliographystyle{colm2026_conference}

\appendix
\section{Omitted details in \Cref{sec:intrp}}
\subsection{Detailed Related Works}\label{sec:related}
We summarize below three lines of existing literature pertinent to our work.
\paragraph{RLHF and LLM Alignment}
Reinforcement learning from human feedback has become the dominant paradigm for aligning large language models with human preferences. Early work demonstrated that reward models trained from human comparisons can guide reinforcement learning policies in complex tasks~\citep{christiano2017deep}, a framework that was later adapted to large language models in systems such as InstructGPT~\citep{ouyang2022training}. Since then, a growing body of work has explored alternative algorithms and formulations for preference optimization and alignment, including preference-based reinforcement learning methods, direct optimization objectives, and improved training pipelines~\citep{stiennon2020learning,bai2022constitutional,rafailov2023direct,yuan2023rrhf,shao2024deepseekmath,yu2025dapo,zheng2025group}. Several studies have further investigated the theoretical foundations, stability properties, and limitations of RLHF-based training~\citep{gao2023scaling,casper2023open}, while others propose scalable alternatives that bypass explicit RL but still optimize preference signals~\citep{ethayarajh2024kto}. Despite differences in algorithms and optimization strategies, these approaches share a common structural assumption: a single reward is assigned to each generated response. In typical RLHF pipelines, a model produces one or several candidate recommendations for a prompt, and a reward model evaluates the whole response, after which policy optimization algorithms such as PPO or related variants update the model to maximize this single sparse reward. While this formulation has proven effective for aligning single responses, it does not naturally capture scenarios where models generate sets of candidate recommendations and the utility depends on the coverage or diversity of the entire set. Such settings arise naturally in many applications, where users select one response from multiple candidates. In contrast to prior work that focuses on sequence-level rewards, our work explicitly models candidate-level reward signals and studies how to allocate reward among candidates in multi-candidate generation using a Shapley-enhanced framework.

\paragraph{Fine-Grained Reward Modeling for LLM Training}
Recent work has moved beyond sequence-level supervision and explored token-level reward assignment for aligning large language models. Instead of assigning a single scalar reward to an entire response, these methods attempt to provide dense token-level reward signals so that different tokens within the same completion can receive different optimization pressures. Representative examples include ABC~\citep{chan2024dense}, which redistributes sequence-level reward to tokens using model attention, TDPO~\citep{zeng2024token}, TLCR~\citep{yoon2024tlcr}, Inverse-Q*~\citep{xia2024inverse}, RTO~\citep{zhong2024dpo}, RED~\citep{li2025red}, T-REG~\citep{zhou2025t}, AlignDistil~\citep{zhang2025aligndistil}, Q-RM~\citep{chen2025discriminative} and TGDPO~\citep{zhu2025tgdpo}. Although these methods provide denser supervision than standard sequence-level RLHF, they also introduce several limitations. In particular, token-level reward assignment often requires additional reward models, auxiliary discriminators, reward redistribution modules, or Monte Carlo estimation, which can significantly increase training complexity. Moreover, many approaches rely on heuristic decompositions of a holistic response reward, such as attention redistribution, self-generated token labels, or induced token-level surrogates, rather than a uniquely justified allocation rule.
% Although these methods provide denser supervision than standard sequence-level RLHF, they also introduce important limitations. First, token-level credit assignment is often computationally expensive, requiring additional reward models, auxiliary discriminators, redistribution modules, Monte Carlo estimation, or token-level regularization objectives. Second, many methods rely on heuristic decompositions of a holistic response reward, for example, via attention redistribution, self-generated token labels, or induced token-level surrogates, rather than a uniquely justified allocation rule. Third, token-level reward is inherently order-sensitive: swapping token positions typically changes semantics, so token attribution does not naturally satisfy the permutation-based fairness principles that motivate Shapley-enhanced reward allocation. In contrast, our work studies a setting where reward is defined over a set of candidates and is invariant to the order of candidates within the set, which makes candidate-level Shapley allocation both better aligned with the task structure and more principled computationally.

\paragraph{Shapley Value in Machine Learning}
Shapley value has been widely adopted in machine learning as a principled tool for contribution attribution, incentive design, and data valuation. A major line of work studies data valuation, beginning with Data Shapley~\citep{ghorbani2019data} and later scalable variants such as \citet{wang2024data}, which reduces the need for repeated retraining by estimating attribution during a single run. Closely related ideas have been used for collaborative and economic allocation, including \citet{sim2020collaborative,schoch2022cs,wang2024economic}. More recently, methods based on Shapley value have also appeared in LLM-related settings, including SCAR~\citep{cao2025scar}, which redistributes a sequence-level reward to tokens or spans for RLHF, as well as Shapley value attribution for prompts, context tokens, and prompt interpretation, such as \citet{liu2023prompt,horovicz-goldshmidt-2024-tokenshap,xiao-etal-2025-tokenshapley}. Shapley ideas have also been applied to LLM fine-tuning data selection and valuation, for example, in \citet{schoch2023data} and recent work on efficient Shapley approximation for LLM fine-tuning~\citep{tamine2025data}. 
However, applying Shapley within a single response introduces an inherent mismatch with the cooperative-game formulation: token- or sentence-level attribution is fundamentally order-sensitive, since swapping tokens or sentences typically changes the semantics of the response. As a result, the "players" in these formulations are not permutation-invariant, which weakens the theoretical justification of Shapley allocation. In contrast, our work studies candidate-level Shapley reward allocation over unordered candidate sets, where the reward is naturally invariant to candidate ordering. This allows our formulation to better align with the cooperative-game motivation of Shapley values while remaining computationally efficient.
% Applying Shapley within a single response creates two limitations for our setting. First, such methods are often computationally heavy and rely on Monte Carlo estimation, surrogate approximations, or other heuristics. Second, token- or sentence-level attribution inside one response is inherently order-sensitive: swapping tokens or sentences typically changes semantics, so the ``players'' are not permutation-invariant in the sense assumed by cooperative-game Shapley allocation. Our work instead studies candidate-level Shapley reward allocation over unordered candidate sets, which better matches the structure of multi-candidate recommendation and preserves the original cooperative-game motivation of Shapley value.

\section{Omitted Proofs in \Cref{sec:theory}}\label{sec:prop}
We begin by reviewing several fundamental properties of the Shapley value, including efficiency, symmetry, additivity, and the null player property.
\begin{itemize}
    \item \emph{Efficiency.} The sum of the Shapley values of all candidates equals the value of the total output, so that all the gain is distributed among the candidates:
    \[
    \sum_{j=1}^K\phi^j=R(o).
    \]
    \item \emph{Symmetry.} If $c^i$ and $c^j$ are two candidates which are equivalent in the sense that $R(S\cup\{c^i\})=R(S\cup\{c^j\})$ for every subset $S$ of  which contains neither $c^i$ nor $c^j$, then $\phi^i=\phi^j$.
    \item \emph{Additivity.} If two cooperative games described by reward functions $R_1$ and $R_2$ are combined, then the distributed rewards should correspond to the rewards derived from $R_1$ and the rewards derived from $R_2$:
    \[
    \phi^i(R_1+R_2)=\phi^i(R_1)+\phi^i(R_2).
    \]
    \item \emph{Null player.} The Shapley value $\phi^i$ of a null player (candidate) $c^i$ in a cooperative game is zero. A candidate $c^i$ is null if $R(S\cup c^i)=R(S)$ for all sets $S$ that do not contain $c^i$. 
\end{itemize}
Efficiency ensures that ShapE-GRPO forms a valid reward allocation rule matching the overall reward. Symmetry guarantees that the reward allocation is independent of the ordering of candidates. Additivity reflects additive rewards. For instance, multiple user clicks on recommended items yield proportional gains in reward. Finally, the null player property implies that incorrect candidates should receive no reward, preventing free-riding behavior.
\subsection{Proof of \Cref{prop:unbias}}
Since we adopt the same $\text{mean}(R)$ and $\text{std}(R)$ as in original GRPO, we only need to prove 
\[
\sum_{t=1}^{|o_i|} R^{\text{ShapE}}(o_{i,t})=|o_i|R(o_i).
\]
We use $|w_i|$ and $|c_i^j|$ to denote the length of the reasoning part and the $j$-th candidate in the $i$-th response. Since all candidates have equal lengths, we know that
\[
\sum_{t=1}^{|o_i|} R^{\text{ShapE}}(o_{i,t})=|w|R(o_i)+\sum_{j=1}^K|c_i^j|K\phi_i^j=|w|R(o_i)+|c_i^1|K\sum_{j=1}^K\phi_i^j.
\]
Due to the efficiency property of the Shapley value, we know that 
\[
\sum_{j=1}^K\phi_i^j=R(o_i).
\]
Therefore, it holds that
\[
\sum_{t=1}^{|o_i|} R^{\text{ShapE}}(o_{i,t})=(|w|+K|c_i^1|)R(o_i)=(|w|+\sum_{j=1}^K|c_i^j|)R(o_i)=|o_i|R(o_i),
\]
which finishes our proof.

\subsection{Proof of \Cref{prop:pos}}
When $|G|>1$, since $R(\cdot)$ is a non-negative function, we know that 
\[\text{mean}(R)\ge 0.
\]
Recall that \[
A_{i,t}^{\text{ShapE}} = \frac{ R^{\text{ShapE}}(o_{i,t})-\text{mean}(R)}{\text{std}(R)}.
\]
For any token corresponding to an incorrect candidate, i.e., $o_{i,t}\in c_i^j$ with $R(c_i^j)=0$, we know that
\[
R^{\text{ShapE}}(o_{i,t})=K\phi_i^j.
\]
Since for any set $S$, we have 
\[
R(S\cup \{c_i^j\})=\max_{c_i^k\in S\cup \{c_i^j\}}R(c_i^k)=\max_{c_i^k\in S}R(c_i^k)=R(S).
\]
Then, the definition of Shapley value immediately yields that 
\[
R^{\text{ShapE}}(o_{i,t})=\sum_{S}
\frac{|S|!(K-|S|-1)!}{K!}
\Big(
R(S \cup \{c^j_i\}) - R(S)
\Big)=K\phi_i^j=K*0=0.
\]
When $|G|=1$, we adopt the widely-used resolution by setting $\text{mean}(R)=0$~\citep{sheng2024hybridflow}. Therefore, together with the above case, it holds that
\[
A_{i,t}^{\text{ShapE}} \le  \frac{ R^{\text{ShapE}}(o_{i,t})}{\text{std}(R)}=0,
\]
when $o_{i,t}\in c_i^j$ and $R(c_i^j)=0$.

\subsection{Proof of \Cref{thm:sort}}
We sort the rewards of the candidates as 
\[
R(c_i^{(1)}) \ge R(c_i^{(2)}) \ge \cdots\ge R(c_i^{(K-1)}) \ge R(c_i^{(K)}).
\]
Recall that the reward of the whole output is $R(o_i)=\max_j R(c_i^{(j)})$. We use $[j]$ to denote $\{1,2,\dots,j\}$ for simplicity and assume $R(c_i^{(K+1)})=0$. It then holds that
\[
R(c_i^{(j)})=\sum_{k=j}^K R(c_i^{(k)})-R(c_i^{(k+1)}).
\]
We define \[
k_i^\star(S) = \arg\min_{j:\, c_i^{(j)} \in S} c_i^{(j)},
\]
and 
\[
U_i^j(S) =
\begin{cases}
1, & \text{if there exists $k\in[j]$ such that $c_i^{(k)}\in S$}, \\
0, & \text{otherwise} .
\end{cases}
\]
Therefore, it holds that
\[
R(S)=R(c_i^{(k_i^\star(S))})=\sum_{k=1}^K (R(c_i^{(k)})-R(c_i^{(k+1)}))U_i^k(S).
\]
Due to the additivity of the Shapley value, we know that
\[
\phi_i^{(j)}(R)=\sum_{k=1}^K (R(c_i^{(k)})-R(c_i^{(k+1)}))\phi_i^{(j)}(U_i^k).
\]
Here $\phi_i^{(j)}$ is the Shapley value of the candidate with the $j$-th largest reward. 

For $\phi_i^{(j)}(U_i^k)$, it holds that
\[
\phi_i^{(j)}(U_i^k) =
\begin{cases}
0, & \text{if }j> k, \\
\frac{1}{k}, & \text{otherwise} .
\end{cases}
\]
The second case holds because 
\[
\sum_j \phi_i^{(j)}(U_i^k)=1,
\]
and the symmetry property of the Shapley value, that is, any $j\le k$ plays the same role for $U_i^k$.

We hence have 
\[
\phi_i^{(j)}=\sum_{k=j}^K \frac{R(c_i^{(k)})-R(c_i^{(k+1)})}{k}.
\]
Using the definition of $R^{\text{ShapE}}(\cdot)$, we know that for $o_{i,t}\in c_i^{(j)}$, it holds that
\[
R^{\text{ShapE}}(o_{i,t})=K\phi_i^{(j)}=\sum_{k=j}^K \frac{K(R(c_i^{(k)})-R(c_i^{(k+1)}))}{k},
\]
which ends the proof.

For the binary case, since our reward function is permutation-invariant, all candidates with the same reward will have the same Shapley value. When $m=0$, it is trivial that all candidates will have Shapley value zero as $R(o_i)=0$. When $m>0$, note that $\sum_j\phi_i^j=R(o_i)=1$. We know that $\phi_i^j=\frac{1}{m}$ when $R(c_i^j)=1$ due to the symmetry property and zero otherwise due to the null player property.
By rescaling by $K$, it yields that
\[
R^{\text{ShapE}} (o_{i,t}) =
\begin{cases}
\frac{K}{|\{j:\, R(c_i^j)=1\}|}, & \text{if }o_{i,t}\in c_i^j \text{ and } R(c_i^j)=1, \\
0, & \text{if }o_{i,t}\in c_i^j \text{ and } R(c_i^j)=0 .
\end{cases}
\]
We remark that $R(\cdot)$ can be any general reward function, even negative, in our proof for \Cref{thm:sort}, demonstrating that our ShapE-GRPO framework can accommodate a wide range of endogenous rewards.

\subsection{Proof of \Cref{corr:flop}}
From the proof of \Cref{thm:sort}, we know that we can utilize \Cref{eq:order} to calculate $R^{\text{ShapE}}(\cdot)$ for every token. For the sorting part, the extra time complexity is $\cO(K\log K)$. For every candidate, the compute overhead using \Cref{eq:order} is $\cO(K)$. Since we have $K$ candidates, the extra overhead is bounded by $\cO(K^2)$. All remaining computations are already part of the standard GRPO procedure.

For the binary case, we only need to derive $m$ in $\cO(K)$ extra overhead. Then, allocating $\frac{K}{m}$ to correct candidates needs another $\cO(K)$ overhead. Therefore, the total extra floating-point operations can be reduced to $\cO(K)$.

Therefore, in the LLM-as-a-recommender setting, incorporating candidate-level dense reward signals incurs only polynomial additional computation, which is negligible in modern reinforcement learning frameworks.
\section{Omitted Experimental Details}\label{app:exp}
\subsection{Dataset Construction}
\begin{figure}[!ht]
\centering
\begin{subfigure}{0.49\linewidth}
\centering
\includegraphics[width=\linewidth]{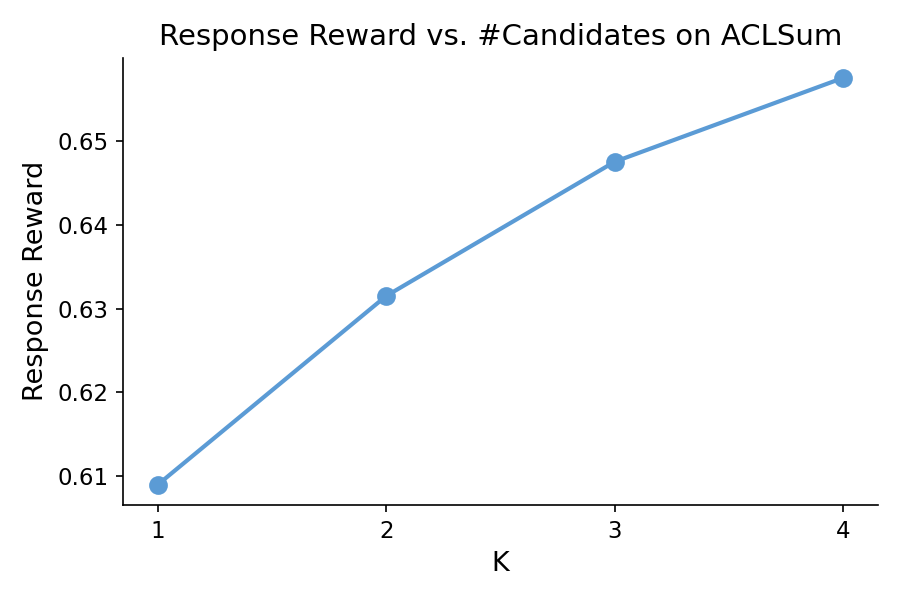}
% \caption{GRPO}
\end{subfigure}
\hfill
\begin{subfigure}{0.49\linewidth}
\centering
\includegraphics[width=\linewidth]{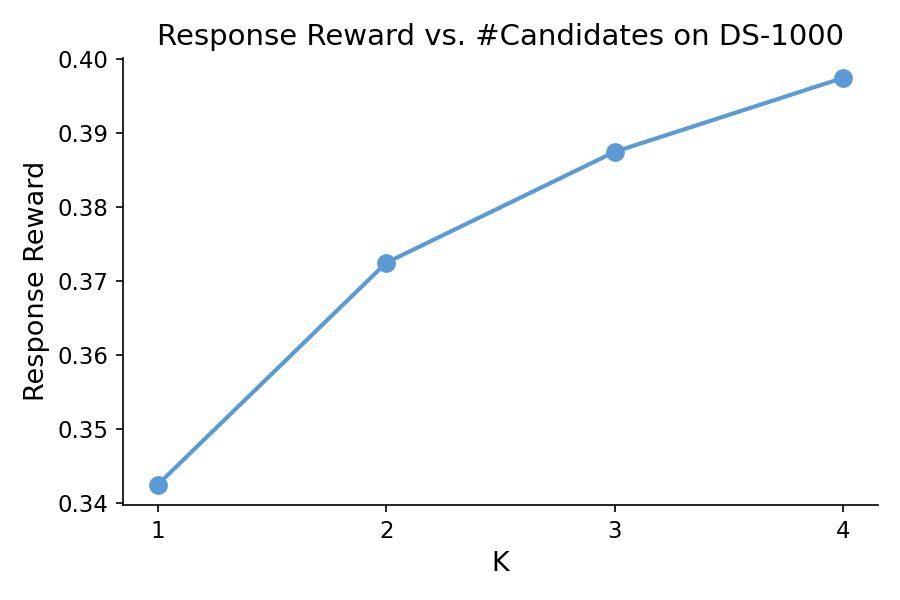}
% \caption{ShapE-GRPO}
\end{subfigure}
\begin{subfigure}{0.49\linewidth}
\centering
\includegraphics[width=\linewidth]{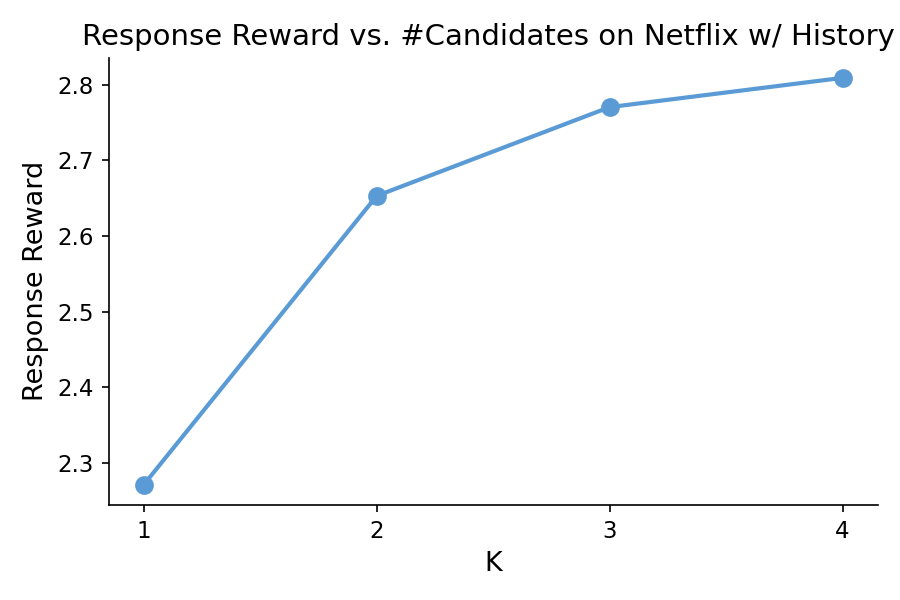}
% \caption{GRPO}
\end{subfigure}
\hfill
\begin{subfigure}{0.49\linewidth}
\centering
\includegraphics[width=\linewidth]{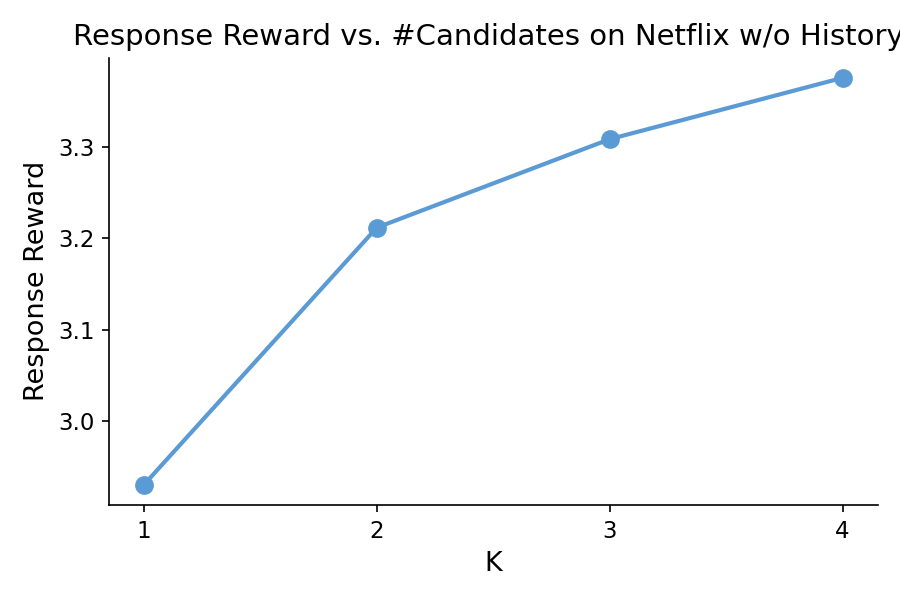}
% \caption{ShapE-GRPO}
\end{subfigure}

\caption{Response reward as a function of the number of candidates across different datasets using \textsc{Qwen3-8b}.}
\label{fig:candidates}
\end{figure}

For all datasets, we require the LLM to generate $K = 4$ candidates. As shown in \Cref{fig:candidates}, the average response reward exhibits diminishing returns as the number of candidates $K$ increases.
Here, we use the first $k$ candidates among all $K$ candidates to simulate the setting where only $k < K$ candidates are generated.
To balance performance and computational cost, we finally choose $K = 4$.

\paragraph{Summarization. }
For the summarization task, we use the ACLSum dataset.
We combine the original training and validation splits
to form the training set, resulting in 150 samples.
The original test split is used as the evaluation set,
which contains 100 samples. For each paper, we ask the LLM to generate four candidate summaries.

For each document, the model is asked to generate a
summary. We adopt an LLM-as-a-judge evaluation protocol
and use \textsc{GPT-5.2} to score the generated summaries
based on the annotated aspects provided in the dataset,
namely challenge, approach, and outcome.
For each aspect, the judge assigns a score from 0 to 3 according to whether the generated summary covers the
corresponding content and the quality of the description.
An additional score of 1 is given to reflect the overall
quality of the summary. The final reward is normalized
to the range $[0,1]$.

\paragraph{Code Suggestion. }
For the code suggestion task, we use the DS-1000 benchmark. For each programming problem, we ask the LLM to generate four candidate solutions, mimicking realistic coding assistance scenarios where a problem may admit multiple valid solution strategies (e.g., Bubble Sort or Quick Sort). As long as one candidate solution is correct, the user can pass the test cases and solve the problem. Therefore, we adopt a binary reward setting in this task. We randomly sample 600 problems to form the training set and use the remaining 400 problems as the evaluation set.

For evaluation, we execute the generated code in a fixed
Python environment. We report the core package versions
needed for our DS-1000 setup:
\texttt{python=3.10},
\texttt{numpy=1.26.4},
\texttt{pandas=1.5.3},
\texttt{scikit-learn=1.4.0},
\texttt{scipy=1.12.0},
\texttt{matplotlib=3.8.4},
\texttt{seaborn=0.13.2},
\texttt{torch=2.2.0}, and
\texttt{tensorflow=2.16.1}.
These packages cover the main dependencies used in our
evaluation pipeline while avoiding unnecessary environment complexity.

\paragraph{Movie Recommendation. }
For the Netflix recommendation task, we construct a dataset based on the Netflix Prize data. We set the number of candidates to $K=4$. We first randomly sample 1000 user data points to form the training pool. For each user, we use their ratings for movies released before 2005 as the user history, and treat movies released in 2005 as the candidate set. In total, there are 512 candidate movies released in 2005.
If a user has not watched any movie released in 2005,
we remove that user from the dataset. To control the
prompt length, we retain at most 800 historical movie
ratings for each user. For the dataset without user history, we simply remove the history information. After filtering, the final training set contains 796 data points.
We construct the evaluation set using the same procedure,
resulting in 775 evaluation instances.

For the reward signal, if the movie recommended by the LLM has been watched by the user and a rating is available, we use the user's rating as the reward. If the user has not watched the recommended movie, we interpret this as the user not accepting the recommendation and assign a reward of $0$.

\subsection{Prompt Templates}
In this section, we list the prompt templates used for the four
datasets and for the LLM-as-a-judge evaluation protocol. Since LLMs have strong instruction-following abilities, we can simply extract candidates from structured outputs.

\paragraph{ACLSum Summarization Prompt.}

\begin{Verbatim}[breaklines=true, breaksymbol=]
You are an expert scientific paper summarizer.

You are given the full text of a scientific paper. Your goal is to produce 4 DISTINCT concise summaries of the paper, each covering the challenge, approach, and outcome.

Paper:
{document}

Your job:
Provide EXACTLY 4 DISTINCT summaries that accurately capture the key information in the paper (challenge, approach, and outcome).

CRITICAL RULES (MUST FOLLOW):
- Each summary MUST be wrapped in <summary> </summary> tags.
- Each summary should be concise (2-4 sentences), covering the challenge addressed, the approach proposed, and the outcome/results.
- Do NOT output any extra text besides the required output format.
- We will use the best summary among the 4 as your final score.
- OUTPUT FORMAT (STRICT):
    "Reasoning: " followed by a short paragraph (one or two sentences). \\n
    "Summaries:" \\n
    1. <summary> Your first summary </summary>\\n
    2. <summary> Your second summary </summary>\\n
    3. <summary> Your third summary </summary>\\n
    4. <summary> Your fourth summary </summary>\\n

EXAMPLE OUTPUT:
Reasoning: I provide four distinct summaries highlighting different aspects of the paper about dependency-based answer ranking for QA.
Summaries:
1. <summary> This paper addresses the challenge of ranking answer candidates for QA when NER produces large candidate sets. The authors propose a statistical method using Dynamic Time Warping on dependency relation paths. Their approach outperforms state-of-the-art methods by up to 20%. </summary>
2. <summary> The authors tackle limitations of surface-level ranking in QA systems by introducing a novel dependency path correlation method. Experiments show significant improvements, especially on harder questions. </summary>
3. <summary> To improve answer ranking in QA, this work uses Dynamic Time Warping to compute correlations between dependency relation paths. The method achieves up to 20% improvement over existing syntactic approaches. </summary>
4. <summary> This paper proposes a statistical ranking approach for QA that leverages dependency structures via DTW. Results demonstrate strong performance gains, particularly when NER-based candidate sets are large. </summary>
\end{Verbatim}

\paragraph{DS-1000 Code Suggestion Prompt.}

\begin{Verbatim}[breaklines=true, breaksymbol=]
You are a Python data processing expert.

You are given a programming task. The problem statement includes any starter code and imports.
Your goal is to produce 4 DISTINCT correct solutions.

Problem:
{problem_description}

Your job:
Provide EXACTLY 4 DISTINCT valid Python programs that correctly solve the problem.

CRITICAL RULES (MUST FOLLOW):
- The 4 solutions MUST be meaningfully different in implementation strategy (not just renaming variables).
- Do NOT output any extra text besides the required output format.
- OUTPUT FORMAT (STRICT):
    "Reasoning: " followed by a short paragraph (one or two sentences). \n
    "Solutions:" \n
    1. <code> python_program </code>\n
    2. <code> python_program </code>\n
    3. <code> python_program </code>\n
    4. <code> python_program </code>\n
- Each python_program must be syntactically valid Python code and executable via exec() without raising a SyntaxError.

EXAMPLE OUTPUT:
Reasoning: I provide four distinct correct approaches while ensuring each program defines result and follows the required formatting.
Solutions:
1. <code> import numpy as np\ndf['#1'] = np.roll(df['#1'], shift=1) </code>
2. <code> Another solution  </code>
3. <code> Third Solution </code>
4. <code> Fourth Solution </code>
\end{Verbatim}

\paragraph{Netflix w/ history Movie Recommendation Prompt.}

\begin{Verbatim}[breaklines=true, breaksymbol=]
You are a movie recommendation system.

You are given:
(A) Historical ratings for movies released BEFORE 2005.
(B) Candidate movies released IN OR AFTER 2005.

User {uid} historical ratings:
{hist_lines}

Candidate movies:
{cand_lines}

Your job:
Pick EXACTLY 4 DISTINCT movies from the candidate list that the user will like most.

CRITICAL RULES (MUST FOLLOW):
- ONLY choose from the candidate list.
- DO NOT copy or repeat the candidate list.
- Output EXACTLY 6 lines and NOTHING ELSE:
  Line 1: "Reasoning: " followed by a short paragraph (one or two sentences).
  Line 2: "Recommendations:"
  Lines 3-6: numbered list in this exact format:
    i. <movie_id> | <movie_title> ({year})
- Use each movie_id at most once.

EXAMPLE OUTPUT:
Reasoning: The user tends to rate dramas and thrillers highly, so I prioritize critically acclaimed films in those genres.
Recommendations:
1. 123 | Example Movie (2006)
2. 456 | Another Movie (2007)
3. 789 | Third Movie (2005)
4. 1011 | Fourth Movie (2008)
\end{Verbatim}

\paragraph{Netflix w/o history Movie Recommendation Prompt.}

\begin{Verbatim}[breaklines=true, breaksymbol=]
You are a movie recommendation system.

You are given:
(A) Candidate movies released IN OR AFTER 2005.

Candidate movies:
{cand_lines}

Your job:
Pick EXACTLY 4 DISTINCT movies from the candidate list that the user will like most.

CRITICAL RULES (MUST FOLLOW):
- ONLY choose from the candidate list.
- DO NOT copy or repeat the candidate list.
- Output EXACTLY 6 lines and NOTHING ELSE:
Line 1: "Reasoning: " followed by a short paragraph (one or two sentences).
Line 2: "Recommendations:"
Lines 3-6: numbered list in this exact format:
    i. <movie_id> | <movie_title> ({year})
- Use each movie_id at most once.

EXAMPLE OUTPUT:
Reasoning: The user tends to rate dramas and thrillers highly, so I prioritize critically acclaimed films in those genres.
Recommendations:
1. 123 | Example Movie (2006)
2. 456 | Another Movie (2007)
3. 789 | Third Movie (2005)
4. 1011 | Fourth Movie (2008)
\end{Verbatim}

\paragraph{LLM-as-a-Judge Prompt.}

\begin{Verbatim}[breaklines=true, breaksymbol=]
You are an EXTREMELY STRICT scientific paper summary evaluator. You must judge whether a generated summary faithfully and precisely covers the three key aspects (challenge, approach, outcome) described in the reference ground truth.

IMPORTANT — Be harsh and precise:
- Only give credit when the summary EXPLICITLY and ACCURATELY reflects the specific content in the ground truth.
- Vague, generic, or paraphrased statements that do not convey the SAME specific meaning as the ground truth should receive LOW scores.
- Do NOT give partial credit for tangentially related statements.
- If the summary fabricates details not present in the ground truth, penalize accordingly.

You will be given:
- A **ground truth** with three components: challenge, approach, and outcome.
- A **generated summary** to evaluate.

Scoring rubric (be strict — most summaries should NOT get full marks):

1. **Challenge coverage** (0-3): Does the summary precisely capture the challenge / problem from the ground truth?
   - 3: Precisely and completely captures the challenge with correct specific details (e.g., exact problem, dataset, or limitation mentioned in GT). Reserve this score for near-perfect matches.
   - 2: Captures the core idea of the challenge but misses important specifics or nuances from the GT.
   - 1: Only vaguely or partially related to the challenge; uses generic language that could apply to many papers.
   - 0: Does not mention the challenge, or the stated challenge is incorrect / irrelevant.

2. **Approach coverage** (0-3): Does the summary precisely capture the approach / method from the ground truth?
   - 3: Precisely and completely describes the approach with correct specific details (e.g., exact method name, technique, or key innovation from GT). Reserve this score for near-perfect matches.
   - 2: Captures the general approach but misses key technical specifics or innovations mentioned in GT.
   - 1: Only vaguely or partially related; describes a generic method without the GT's specific details.
   - 0: Does not mention the approach, or the stated approach is incorrect / irrelevant.

3. **Outcome coverage** (0-3): Does the summary precisely capture the outcome / results from the ground truth?
   - 3: Precisely and completely captures the outcome with correct specific details (e.g., exact metrics, improvements, or findings from GT). Reserve this score for near-perfect matches.
   - 2: Captures the general outcome but misses specific numbers, comparisons, or key findings from GT.
   - 1: Only vaguely mentions results; uses generic language like "achieves good performance" without GT specifics.
   - 0: Does not mention the outcome, or the stated outcome is incorrect / irrelevant.

4. **Overall quality** (0-1): Is the summary well-written, coherent, and concise?
   - 1: Well-organized, fluent, and appropriately concise. No hallucinated content.
   - 0.5: Acceptable but has minor issues (verbose, awkward phrasing, or minor inaccuracies).
   - 0: Poorly written, incoherent, excessively verbose/short, or contains hallucinated content.

Total score = challenge_score + approach_score + outcome_score + overall_score  (range: 0-10)

CALIBRATION GUIDELINE:
- A perfect summary that matches all GT details precisely: 9-10
- A good summary that captures main ideas but misses some specifics: 5-7
- A mediocre summary with only vague coverage: 2-4
- A poor or irrelevant summary: 0-2

Return ONLY a JSON object in this exact format (no other text):
{"challenge_score": <float 0-3>, "approach_score": <float 0-3>, "outcome_score": <float 0-3>, "overall_score": <float 0-1>, "score": <float 0-10>, "reason": "<brief explanation>"}
\end{Verbatim}

\subsection{Training Hyperparameters}
We implement both ShapE-GRPO and GRPO using the \texttt{verl}
training framework~\citep{sheng2024hybridflow}. For each prompt, we perform a single rollout that generates four candidates. When training the \textsc{Qwen3-8B} model, we observe
that the model frequently generates long reasoning traces within the \texttt{<think>} and \texttt{</think>} tags, which can cause the
response length to exceed the predefined limit. To control the
length of the reasoning process, we introduce a length penalty on the thinking segment.
We set the target thinking length to 2,048 tokens. If the actual thinking length is $l > 2048$, we apply a penalty proportional to the overflow $(l - 2048)/2048$. For ShapE-GRPO, this penalty is applied at the token level, assigning a penalty of $(l - 2048)/2048$ to each token that exceeds the limit. That is
\[
\text{length penalty}=\frac{\max(l-2048,0)}{2048}.
\]
For GRPO, since the reward signal is sparse and assigned only at the sequence level, we subtract the same penalty
$(l - 2048)/2048$ directly from the overall reward.

We then report the main training hyperparameters used in our experiments. \Cref{tab:training_hparams} summarizes the most important settings. Extending this to larger-scale experiments is an interesting direction.

\begin{table}[!ht]
\centering
% \small
\begin{tabular}{>{\raggedright\arraybackslash}p{0.4\linewidth}
                >{\centering\arraybackslash}p{0.4\linewidth}}
\toprule
Hyperparameter & Value \\
\midrule
Training epochs & 1 \\
Train batch size & 2 \\
Learning rate & 1e-6 \\
Max response length & 4096 \\
Sampling temperature & 0.7 \\
Top-$p$ & 0.9 \\
% Number of samples per prompt & 1 \\
KL coefficient ($\beta$) & 0.01 \\
Clip range ($\epsilon$) & 0.2 \\
\bottomrule
\end{tabular}
\caption{Main training hyperparameters used in our experiments.}
\label{tab:training_hparams}
\end{table}
To demonstrate the effectiveness of ShapE-GRPO, we intentionally avoid extensive hyperparameter tuning for specific models or datasets, and instead use a unified set of training parameters across all experiments. Additional fine-grained tuning could potentially further improve the performance of ShapE-GRPO, which we leave for future work.

\subsection{Additional Results}
To further evaluate the generality of ShapE-GRPO across different large language model architectures, we conduct additional experiments using \textsc{Llama 3.1‑8B‑Instruct}. Given that WTA consistently underperforms traditional GRPO, we focus on the comparison between ShapE-GRPO and GRPO.  We adopt the same training configuration used in the experiments with \textsc{Qwen3‑8B}, including identical hyperparameters, training procedures, and evaluation protocols.

\begin{table}[!ht]
\centering
\small
\begin{tabular}{lccc}
\toprule
Dataset & \textsc{Llama 3.1‑8B‑Instruct} & GRPO & ShapE-GRPO \\
\midrule
ACLSum ($\uparrow$) & 0.664 & 0.707 & \textbf{0.780} \\
DS-1000 ($\uparrow$) & 0.180 & 0.205 & \textbf{0.278} \\
Netflix w/ history ($\uparrow$) & 1.885 & 3.783 & \textbf{4.027} \\
Netflix w/o history ($\uparrow$) & 2.130 & 3.783 & \textbf{4.071} \\
\bottomrule
\end{tabular}
\caption{Performance comparison across four datasets using \textsc{Llama 3.1‑8B‑Instruct}.
ShapE-GRPO consistently improves over both the pretrained
model and the standard GRPO baseline.}
\label{tab:llama}
\end{table}

An interesting observation from \Cref{tab:llama} is that the performance on Netflix with user history remains slightly lower than that of the setting without history, even after training with ShapE-GRPO. This suggests that \textsc{Llama 3.1-8B-Instruct} may not fully utilize the long contextual information contained in user interaction histories. In recommendation tasks, informative signals are often sparse and distributed across long prompts, which can make it challenging for the model to effectively extract the most relevant information during both training and inference. As a result, the potential benefit of additional historical context may not be fully realized.
One possible direction to address this limitation is to integrate long-context optimization techniques~\citep{ge2023context} to improve the effectiveness and efficiency of attention over long sequences. Exploring the interaction between such long-context techniques and ShapE-GRPO constitutes an interesting direction for future work.

Overall, \Cref{tab:llama} shows that ShapE-GRPO consistently improves over both the pretrained model and the standard GRPO baseline across all datasets when applied to \textsc{Llama 3.1-8B-Instruct}. These results further support that the benefits of candidate-level reward allocation are not tied to a specific model architecture, but instead generalize across different LLM families.

\iffalse
\section{Appendix}
You may include other additional sections here.
\fi

\end{document}